\documentclass[11pt,a4paper]{article}
\usepackage[hyperref]{naaclhlt2018}
\usepackage{times}
\usepackage{latexsym}
\usepackage{graphicx}
\usepackage{caption}
\usepackage{url}

\aclfinalcopy 

\title{MaskParse@Deskin at SemEval-2019 Task 1:  Cross-lingual UCCA Semantic Parsing using Recursive Masked Sequence Tagging}

\author{Gabriel Marzinotto$^{1, 2}$ ~~~~~ Johannes Heinecke$^{1}$ ~~~~~ G\'eraldine Damnati$^{1}$  \\
  (1) Orange Labs / Lannion France \\
  (2) Aix Marseille Univ, CNRS, LIS / Marseille France \\
  {\tt \{gabriel.marzinotto,johannes.heinecke,geraldine.damnati\}@orange.com}}

\date{}

\begin{document}
\maketitle
\begin{abstract}
This paper describes our recursive system for SemEval-2019 \textit{ Task  1: Cross-lingual Semantic Parsing with UCCA}. Each recursive step consists of two parts. We first perform semantic parsing using a sequence tagger to estimate the probabilities of the UCCA categories in the sentence. Then, we apply a decoding policy which interprets these probabilities and builds the graph nodes. Parsing is done recursively, we perform a first inference on the sentence to extract the main scenes and links and then we recursively apply our model on the sentence using a masking feature that reflects the decisions made in previous steps. Process continues until the terminal nodes are reached. We choose a standard neural tagger and we focused on our recursive parsing strategy and on the cross lingual transfer problem to develop a robust model for the French language, using only few training samples. 
\end{abstract}

\section{Introduction}
Semantic representation is an essential part of NLP. For this reason, several semantic representation paradigms have been proposed. Among them we find PropBank \cite{palmer2005proposition} and  FrameNet Semantics \cite{baker1998berkeley}, Abstract Meaning Representation (AMR) \cite{banarescu2013abstract}, Universal Decompositional Semantics \cite{white2016universal} and Universal Conceptual Cognitive Annotation (UCCA) \cite{abend2013universal}. These constantly improving representations, along with the advances in semantic parsing, have proven to be beneficial in many NLU tasks such as Question Answering \cite{inproceedings_QA}, text summarization \cite{abstractive_summary}, dialog systems \cite{inproceedings_slu_srl}, information extraction \cite{W13-3820} and machine translation \cite{Liu:2010:SRF:1873781.1873862}.

UCCA is a cross-lingual semantic representation scheme, has demonstrated applicability in  English, French and German (with pilot annotation projects on Czech, Russian and Hebrew). Despite the newness of UCCA, it has proven useful for defining semantic evaluation measures in text-to-text generation and machine translation \cite{birch2016hume}. UCCA represents the semantics of a sentence using directed acyclic graphs (DAGs), where terminal nodes correspond to text tokens, and non-terminal nodes to higher level semantic units. Edges are labelled, indicating the role of a child in the relation to its parent. UCCA parsing is a recent task and since UCCA has several unique properties, adapting syntactic parsers or parsers from other semantic representations is not straight-forward. Current state of the art parser TUPA \cite{hershcovich2017transition} uses a transition based parsing to build UCCA representations.

Building over previous work on FrameNet Semantic Parsing \cite{marzinotto:calor,marzinotto:hal-01731385} we chose to perform UCCA parsing using sequence tagging methods along with a graph decoding policy. To do this we propose a recursive strategy in which we perform a first inference on the sentence to extract the main scenes and links and then we recursively apply our model on the sentence with a masking mechanism at the input in order to feed information about the previous parsing decisions. 

\section{Model}
\label{sec:model}
Our system consists of a sequence tagger that is first applied on the sentence to extract the main scenes and links and then it is recursively applied on the extracted element to build the semantic graph. At each step of the recursion we use a masking mechanism to feed information about the previous stages into the model. In order to convert the sequence labels into nodes of the UCCA graph we also apply a decoding policy at each stage. 

Our tagger is implemented using deep bi-directional GRU (\emph{biGRU}). This simple architecture is frequently used in semantic parsers across different representation paradigms. Besides its flexibility, it is a powerful model, with close to state of the art performance on both PropBank \cite{he2017deep} and FrameNet semantic parsing
\cite{SoinFrameParsing,
marzinotto:hal-01731385}.

More precisely, the model consists of a 4 layer bi-directional Gated Recurrent Unit (GRU) with highway connections \cite{SrivastavaGS15}. Our model uses has a rich set of features including syntactic, morphological, lexical and surface features, which have shown to be useful in language abstracted representations. The list is given below:

\begin{itemize}
\setlength\itemsep{-4pt}
\item Word embeddings of 300 dimensions \footnote{Obtained from https://github.com/facebookresearch/MUSE}.
\item Syntactic dependencies of each token\footnote{\label{note1} Using Universal Dependencies categories. }.  
\item Part-of-speech and morphological features such as gender, number, voice and degree\footnotemark[2].
\item Capitalization and word length encoding. 
\item Prefixes and Suffixes of 2 and 3 characters.
\item A language indicator feature.
\item Boolean indicator of idioms and multi word expression. Detailed in section \ref{sec:expressions}.
\item Masking mechanism, which indicates, for a given node in the graph, the tokens within the span as well as the arc label between the node and its parent. See details in section \ref{sec:masking}.
\end{itemize}

Except for words where we use pre-trained embeddings, we use randomly initialized embedding layers for categorical features.

\begin{figure*}[htbp]
  \centering
  \includegraphics[width=0.95\linewidth]{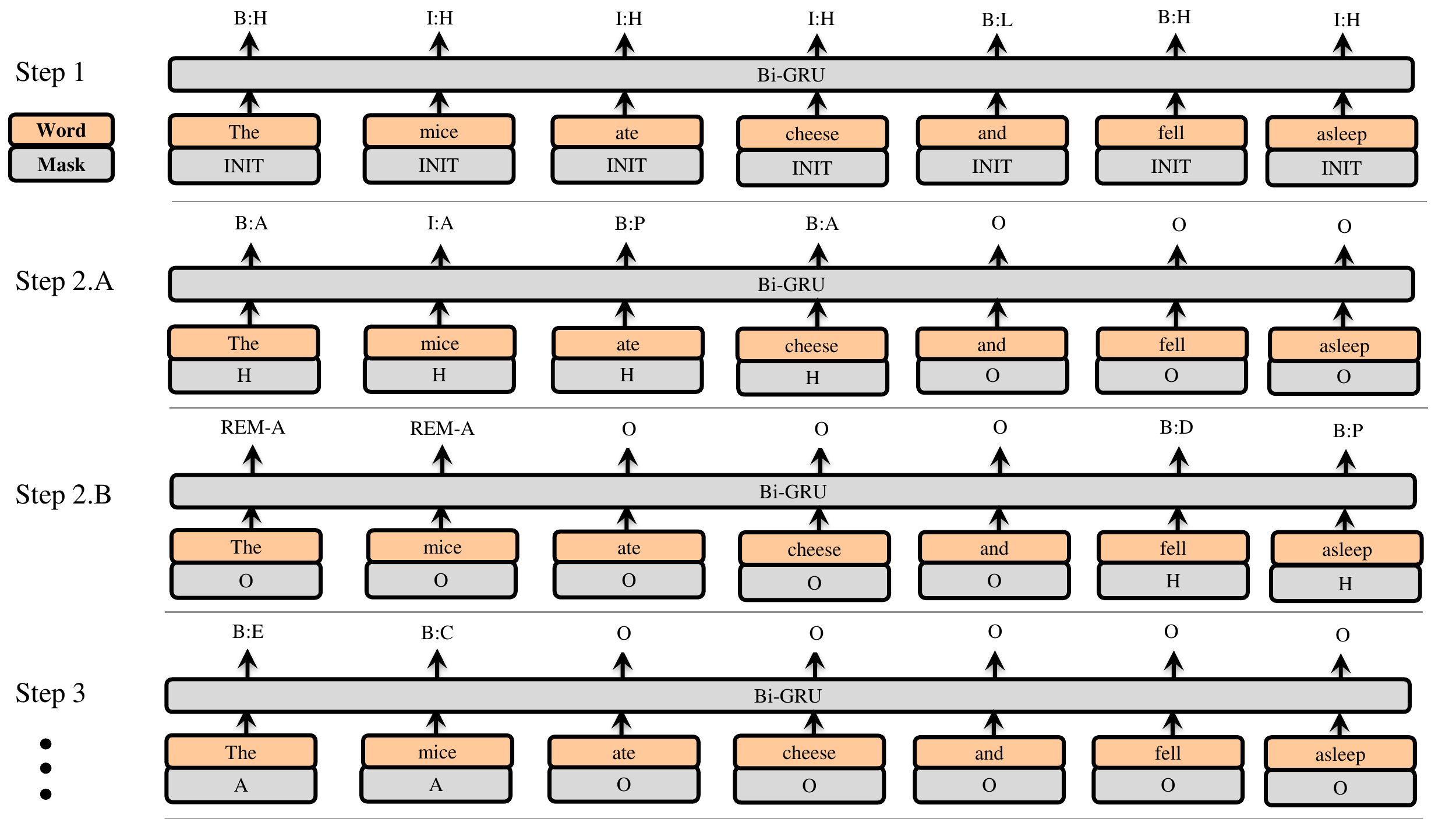}
  \caption{ Masking mechanism through recursive calls. \texttt{Step 1} parses the sentence to extract parallel scenes (H) and links (L). Then \texttt{Steps 2.A 2.B} use a different mask to parse these scenes and extract arguments (A) and processes (P) which will be recursively parsed until terminal nodes are reached.}
  \label{fig:hwlstm}
  \vspace{-4mm}
\end{figure*}  

\subsection{Masking Mechanism}
\label{sec:masking}
We introduce an original masking mechanism in order to feed information about the previous parsing stages into the model. During parsing, we first do an initial inference step to extract the main scenes and links. Then, for each resulting node, we build a new input which is essentially the same, but with a categorical sequence masking feature. For the input tokens in the node span, this feature is equal to the label of the arc between the node and its parent. Outside of the node span, this mask is equal to \texttt{O}. A diagram of this masking process is shown in figure \ref{fig:hwlstm}. The process continues and the model recursively extracts the inner semantic structures (the node's children) in the graph, until the terminal nodes are reached.

To train such a model, we build a new training corpus in which the sentences are repeated several times. More precisely, a sentence appears $N$ times ($N$ being the number of non terminal nodes in the UCCA graph) each one a with different mask.

\subsection{Multi-Task UCCA Objective}
Along with the UCCA-XML graph representations, a simplified tree representation in CoNLL format was also provided.  
Our model combines both representations using a multitask objective with two tasks. \texttt{TASK1} consists in, for a given node and its corresponding mask, predicting the children and their arc labels. \texttt{TASK1} encodes the children spans using a BIO scheme. The \texttt{TASK2} consists in predicting the CoNLL simplified UCCA structure of the sentence. More precisely, \texttt{TASK2} is a sequence tagger that predicts the UCCA-CoNLL function of each token. \texttt{TASK2} is not used for inference purposes. It is only a support that help the model to extract relevant features, allowing it to model the whole sentence even when parsing small pre-terminal nodes.

\subsection{Label Encoding}
We have previously stated that \texttt{TASK1} uses BIO encoded labels to model the structure of the children of each node in the semantic graph. In some rare cases, the BIO encoding scheme is not sufficient to model the interaction between parallel scenes. For example, when we have two parallel scenes and one of them appears as a clause inside the other. In such cases, BIO encoding does not allow to determine whether the last part of the sentence belongs to the first scene or to the clause. Despite this issue, prior experiments testing more complete label encoding schemes (BIEO, BIEOW) showed that BIO outperforms the other schemes on the validation sets.

\subsection{Graph Decoding}
During the decoding phase, we convert the BIO labels into graph nodes. To do so, we add a few constraints to ensure the outputs are feasible UCCA graphs that respect the sentence's structure:
\begin{itemize}
\setlength\itemsep{-4pt}
\item We merge parallel scenes (H) that do not have either a verb or an action noun to the nearest previous scene having one.
\item Within each parallel scene, we force the existence of one and only one \texttt{State} (S) or \texttt{Process} (P) by selecting the token with the highest probability of \texttt{State} or \texttt{Process}.
\item For scenes (H) and arguments (A) we do not allow to split multi word expressions (MWE) and chunks into different graph nodes. If the boundary between two segments lies inside a chunk or MWE segments are merged.
\end{itemize}

\subsection{Remote Edges}
Our approach easily handles remote edges. We consider remote arguments as those detected outside the parent's node span (see \texttt{REM} in Fig.\ref{fig:hwlstm}). Our earlier models showed low recall on remotes. To fix this, we introduced a detection threshold on the output probabilities. This increased the recall at the cost of some precision. The optimal detection threshold was optimized on the validation set.


\section{Data}

\subsection{UCCA Task Data}
In table \ref{tab:data} we show the number of annotations for each language and domain. Our objective is to build a model that generalizes to the French language despite of having only 15 training samples.

\begin{table}[t!]
\centering
\begin{tabular}{lrrr}
  Corpus & Train & Dev & Test \\
  \hline
  English Wiki & 4113 & 514 & 515 \\
  English 20K  & -    & -   & 492 \\
  German  20K  & 5211 & 651 & 652 \\
  French  20K  & 15   & 238 & 239 \\
\end{tabular}
\caption{number of UCCA annotated sentences in the partitions for each language and domain}
\label{tab:data}
\vspace{-4mm}
\end{table}

\begin{table*}[!htbp]
\centering
\begin{tabular}{l|ccc|ccc|ccc|ccc|}
    & \multicolumn{3}{c}{Ours Labeled} & \multicolumn{3}{|c|}{Ours Unlabeled} & \multicolumn{3}{|c|}{TUPA Labeled} & \multicolumn{3}{c|}{TUPA Unlabeled}\\
  \cline{2-13}
   Open Tracks & \multicolumn{1}{|p{0.6cm}|}{\centering Avg \\ F1} & \multicolumn{1}{|p{0.6cm}|}{\centering Prim \\ F1}  & \multicolumn{1}{|p{0.6cm}|}{\centering Rem \\ F1}   & \multicolumn{1}{|p{0.6cm}|}{\centering Avg \\ F1}   & \multicolumn{1}{|p{0.6cm}|}{\centering Prim \\ F1}  & \multicolumn{1}{|p{0.6cm}|}{\centering Rem \\ F1}   & \multicolumn{1}{|p{0.6cm}|}{\centering Avg \\ F1}   & \multicolumn{1}{|p{0.6cm}|}{\centering Prim \\ F1}  & \multicolumn{1}{|p{0.6cm}|}{\centering Rem \\ F1}   & \multicolumn{1}{|p{0.6cm}|}{\centering Avg \\ F1}   & \multicolumn{1}{|p{0.6cm}|}{\centering Prim \\ F1}  & \multicolumn{1}{|p{0.6cm}|}{\centering Rem \\ F1} \\
  \hline
  Dev  English Wiki & \textbf{70.8} & 71.3 & 58.7 & 82.5 & 83.8 & 37.5 & \textbf{74.8} & 75.3 & 51.4 & 86.3 & 87.0 & 51.4 \\
  Dev  German  20K  & \textbf{74.7} & 75.4 & 40.5 & 87.4 & 88.6 & 40.9 & \textbf{79.2} & 79.7 & 58.7 & 90.7 & 91.5 & 59.0 \\
  Dev  French  20K  & \textbf{\underline{63.6}} & 64.4 & 19.0 & 78.9 & 79.6 & 20.5 & \textbf{\underline{51.4}} & 52.3 & 1.6 & 74.9 & 76.2 & 1.6 \\
  \hline
  Test English Wiki & \textbf{68.9} & 69.4 & 42.5    & 82.3 & 83.1 & 42.8 & \textbf{73.5} & 73.9 & 53.5 &     85.1 & 85.7 & 54.3 \\
  Test English 20K  & \textbf{66.6} & 67.7 & 24.6    & 82.0 & 83.4 & 24.9 & \textbf{68.4} & 69.4 & 25.9 & 82.5 & 83.9 & 26.2 \\
  Test German  20K  & \textbf{74.2} & 74.8 & 47.3    & 87.1 & 88.0 & 47.6 & \textbf{79.1} & 79.6 & 59.9 & 90.3 & 91.0 & 60.5\\
  Test French  20K  & \textbf{\underline{65.4}} & 66.6 & 24.3    & 80.9 & 82.5 & 25.8 & \textbf{\underline{48.7}} & 49.6 & 2.4 &	74.0 & 75.3 & 3.2  \\
  \hline
\end{tabular}
\caption{Our model vs TUPA baseline performance for each open track}
\label{tab:results_pt1}
\vspace{-2mm}
\end{table*}

\begin{table*}[!htbp]
\centering
\begin{tabular}{l|c|c|c|c|c|c|c|c|c|c|c|c|c|}
   Tracks & D & C & N & E & F & G & L & H & A & P & U & R & S \\
  \hline
  EN Wiki & 64.3 & 71.4 & 68.5 & 69.6 & 76.7 & 0.0 & 71.4 & 61.3 & 60.0 & 64.0 & 99.7 & 89.2 & 25.1 \\
  EN 20K  & 47.2 & 75.2 & 62.5 & 72.3 & 71.5 & 0.2 & 57.9 & 49.5 & 55.7 & 69.8 & 99.7 & 83.2 & 19.5  \\
  DE 20K  & 69.4 & 83.8 & 57.7 & 80.5 & 83.8 & 59.2 & 68.4 & 62.2 & 67.5 & 68.9 & 97.1 & 86.9 & 25.9 \\
  FR 20K  & 46.1 & 76.0 & 58.9 & 71.2 & 53.3 & 4.8 & 59.4 & 50.4 & 52.8 & 67.6 & 99.6 & 83.5 & 16.9  \\
  \hline
\end{tabular}
\caption{Our model's Fine-grained F1 by label on Test Open Tracks }
\label{tab:results_pt2}
\vspace{-4mm}
\end{table*}

When we analyse data in details we observe that there are several tokenization errors. Specially in the French corpus. These errors propagate to the POS tagging and dependency parsing as well. For this reason, we retokenized and parsed all the corpus using a enriched version of UDpipe that we trained ourselves \cite{udpipe:2017} using the Treebanks from Universial Dependencies\footnote{\url{https://universaldependencies.org/}}. For French we enriched the Treebank with XPOS from our lexicon. Finally, since tokenization is pre-established in the UCCA corpus we projected the improved POS and dependency parsing into the original tokenization of the task. 

\subsection{Supplementary lexicon}
\label{sec:expressions}
We observed that a major difficulty in UCCA parsing is analyzing idioms and phrases. The unawareness about these expressions, which are mostly used as links between scenes, mislead the model during the early stages of the inference and errors get propagated through the graph. To boost the performance of our model when detecting links and parallel scenes we developed an internal list with about 500 expression for each language. These lists include prepositional, adverbial and conjunctive expressions and are used to compute Boolean features indicating the words in the sentence which are part of an expression.

\subsection{Multilingual Training}

This model uses multilingual word embeddings trained using fastText \cite{bojanowski2017enriching} and aligned using MUSE \cite{conneau2017word}. This is done in order to ease cross-lingual training.
In prior experiments we introduced an adversarial objective similar to \cite{D17-1302,  naacl-advlearning} to build a language independent representation. However, the language imbalance on the training data did not allow us to take advantage from this technique. Hence, we simply merged training data from different languages. 

\section{Experiments}

We focus on obtaining the model that best generalizes on the French language. We trained our model for 50 epochs and we selected the best one on the validation set. In our experiments we did not use any product of experts or bagging technique and we did not run any hyper parameter optimization.  

We trained several models building different training corpora composed of different language combinations. We obtained our best model using the training data for all the languages. This model \texttt{FR+DE+EN} achieved 63.6\% avg. F1 on the French validation set. Compared to 63.1\% for \texttt{FR+DE}, 62.9\% for \texttt{FR+EN} and 50.8\% for only \texttt{FR}.

\subsection{Main Results}

In Table \ref{tab:results_pt1} we provide the performance of our model for all the open tracks and we provide the results for TUPA baseline in order to establish a comparison. Our model finishes 4th in the French Open Track with an average F1 score of 65.4\%, very close to the 3rd place which had a 65.6\% F1. For languages with larger training corpus, our model did not outperform the monolingual TUPA. 

\subsection{Error Analysis}

In Table \ref{tab:results_pt2} we give the performance by arc type. We observe that the main performance bottleneck is in the parallel scene segmentation (H). Due to our recursive parsing approach, this kind of error is particularly harmful to the model performance, because scene segmentation errors at the early steps of the parsing may induce errors in the rest of the graph. To assert this, we used the validation set to compare the performance of the mono scene sentences (with no potential scene segmentation problems) with the multi scene sentences. For the French track we obtained 67.2\% avg. F1 on the 114 mono scene sentences compared to 61.9\% avg. F1 on the 124 multi scene sentences.

\section{Conclusions}
We described an original approach to recursively build the UCCA semantic graph using a sequence tagger along with a masking mechanism and a decoding policy. Even though this approach did not yield the best results in the UCCA task, we believe that our original recursive, mask-based parsing can be helpful in low resource languages. Moreover, we believe that this model could be further improved by introducing a global criterion and by performing further hyper parameter tuning. 


\bibliography{semeval2018}
\bibliographystyle{acl_natbib}

\end{document}